\documentclass[a4paper,12pt]{article}
\usepackage{paralist}
\usepackage{sidecap}
\usepackage{graphicx}
\usepackage{caption}
\usepackage{subcaption}
\usepackage{wrapfig}
\usepackage[table]{xcolor}
\usepackage{pifont}
\usepackage{verbatim}
\usepackage{hyperref}

\begin{document}
  \title{A Detailed Rubric for Motion Segmentation}
  \author{Pia Bideau and Erik Learned-Miller}
  \date{}
  \maketitle
  \begin{abstract}
  Motion segmentation is currently an active area of research in computer Vision. The task of comparing different methods of motion segmentation is complicated by the fact that researchers may use subtly different definitions of the problem. Questions such as "Which objects are moving?", "What is background?", and "How can we use motion of the camera to segment objects, whether they are static or moving?" are clearly related to each other, but lead to different algorithms, and imply different versions of the ground truth.
  
  This report has two goals. The first is to offer a precise definition of motion segmentation so that the intent of an algorithm is as well-defined as possible. The second is to report on new versions of three previously existing data sets that are compatible with this definition. We hope that this more detailed definition, and the three data sets that go with it, will allow more meaningful comparisons of certain motion segmentation methods. 
  \end{abstract}
  
  \section{Introduction}
  The general idea of using motion of the camera and motions of objects to break the world into segments is a fundamental area of research in computer vision. There are many basic approaches to this idea including the following.
  \begin{itemize}
  \item With a static camera, one can analyze parts of the image that change from frame to frame as "foreground". This technique often goes by the name of {\em background subtraction} or {\em background modeling}. 
  \item When a camera is moving, then different parts of the world move in different ways within the image, whether they are actually in motion themselves or not. Reconstructing the geometry of a static world using the motion of a camera is typically called {\em structure from motion}, and while this problem is usually considered distinct from {\em motion segmentation}, we can see elements of structure from motion that appear in works on motion segmentation.
  \item When a camera is moving and elements of the environment are also moving, it can be challenging to separate the world into parts. This is the general area of {\em motion segmentation}, but there are many different ways to define it. For example, should an object that is moving in one frame and still in the next frame be segmented by such an algorithm in the still frame? Or, should an object like a person, part of which is moving, and part of which is not moving be fully segmented, or should we only segment the piece of the object which is moving? Different answers to these questions lead to different ground truths, and hence to different algorithm goals.
  \end{itemize}
  
  There is, of course, no single correct definition of {\em motion segmentation}. However, despite the lack of a universal definition, it is still important to have good agreement between the definition one has accepted as a research problem and the ground truth of the data set that it will be evaluated on. 
  In our research, we have identified the need for new versions of the ground truth of several published data sets, based upon our working definition of motion segmentation. In this report, we give a highly detailed definition and discussion of our definition of motion segmentation, and offer new versions of the ground truth of three pre-existing data sets. These data sets are
  \begin{itemize}
  \item FBMS-59,
  \item Complex Background,
  \item Camouflaged Animals.
  \end{itemize}
Of course, the original versions of these data sets will still be useful for the original purposes put forward by the authors, but these new versions will allow more accurate comparisons for the specific problem definition we are putting forward.

In addition to the topics discussed above, there is also the cross-cutting area of {\em video segmentation}. While this does not mention motion explicitly, it is clearly closely related to motion segmentation since moving objects are natural to segment in videos. A clear difference is that in video segmentation, it is reasonable to segment objects that are not moving, like a statue, while in motion segmentation, this would not fit the definition. Thus, while these problems are clearly related on the surface, they lead to different ground truths, and thus it is important to be clear on which problem one is addressing.
  
The new versions of the ground truth accordingly to our problem definition can be downloaded \url{http://vis-www.cs.umass.edu/motionSegmentation/reportDef/report.html}.
   
  \section{Data Sets}
  In this section we will briefly review the most popular data sets which are widely used for motion segmentation and closely related problems. The intention of the provided ground truth might differ quite a lot even though the problem seams to be similar.
  Often the selected ground truth to evaluate a specific approach doesn't match the addressed problem which is intended to solve exactly. If the addressed problem doesn't match the intention of the ground truth used for evaluation an accurate evaluation gets impossible. Therefore being aware of the problem definition, what one tries to solve, and the specific goal of the ground truth which is used for evaluation is critical.
  
\vspace{2ex}
  \textbf{Freiburg-Berkeley Motion Segmentation Dataset (FBMS-59)\cite{ochs2014segmentation}} The data set consists of 59 video sequences, containing 12 traffic scenes, 13 short extracts of the series ”Mrs. Marple”, 7 indoor scenes showing cats or rabbits and 32 other outdoor scenes showing mostly horses, dogs and people. The traffic scenes are a subset of the large Hopkins 155 data set with new annotations. 720 frames are annotated. The typical frame size varies between 350x288 pixel and 960x540 pixel.
  FBMS-59 is split into a training set and a test set. Typical challenges appear in both sets. 
  The FBMS-59 addresses the problem of segmenting moving objects. Their understanding of moving object segmentation is based on the Gestalt principle of "common fate" \cite{koffka2013principles}. Pixels which share the same motion are grouped together. This approach incorporates principles of motion perception studied in psychology and cover fundamental problems, however it still comes with a lot of uncertainties for practical applications in computer vision. What exactly does ”same motion” mean? Does that mean same motion direction? 
  The provided ground truth shows a depth dependent segmentation. The ground truth of the videos marple2 and marple10 shows segmented walls in the foreground labeled as moving objects. This suggests a depth dependent understanding of motion segmentation. 
  This is not what matches the definition of motion segmentation provided in Section ~\ref{sec:MoSegDef}. We don't want to segment walls, since they are not moving in 3D.
  The given ground truth requires a non-causal system, since an object is segmented as moving even when it is actually not moving but will move in future or moved in the past. This does not correspond to mechanisms of the causal human vision system and thus human motion perception. This non causal approach is not solvable in real time. The entire video has to be analyzed before segmenting the first frame. However a consistent object segmentation over time might be of importance for some applications for example in computer graphics and video editing. However this is a slightly different problem;
   Segmenting an object which has moved at some point in the video is not considered as the fundamental motion segmentation problem.
  
  This database can be viewed and downloaded at \url{http://lmb.informatik.uni-freiburg. de/resources/datasets/}
   
 %\begin{comment}
 \begin{figure}
     \centering
     \begin{subfigure}[t] {0.15\textwidth}
     	 \centering
         \includegraphics[height=1.7cm]{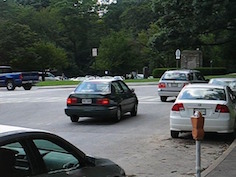}
         \includegraphics[height=1.7cm]{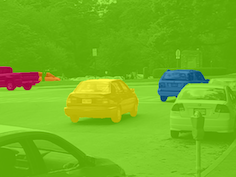}
         \caption{\small cars2}
         \label{fig:gull}
     \end{subfigure}
     ~ %add desired spacing between images, e. g. ~, \quad, \qquad, \hfill etc. 
     \begin{subfigure}[t] {0.2\textwidth}
      \centering
         \includegraphics[height=1.7cm]{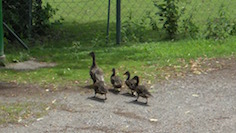}
         \includegraphics[height=1.7cm]{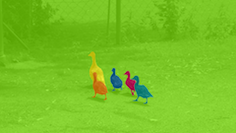}
         \caption{\small ducks01}
         \label{fig:tiger}
     \end{subfigure}
     ~ %add desired spacing between images, e. g. ~, \quad, \qquad, \hfill etc. 
     \begin{subfigure}[t] {0.16\textwidth}
     	 \centering
         \includegraphics[height=1.7cm]{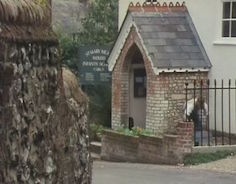}
         \includegraphics[height=1.7cm]{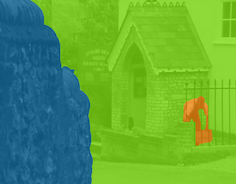}
         \caption{\small marple10}
         \label{fig:mouse}
     \end{subfigure}
     \begin{subfigure}[t] {0.18\textwidth}
     	  \centering
          \includegraphics[height=1.7cm]{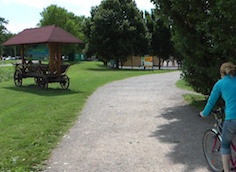}
          \includegraphics[height=1.7cm]{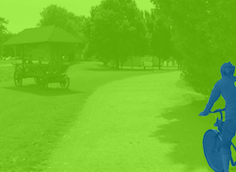}
          \caption{\small people05}
          \label{fig:mouse}
      \end{subfigure}
      \begin{subfigure}[t] {0.18\textwidth}
      	   \centering
           \includegraphics[height=1.7cm]{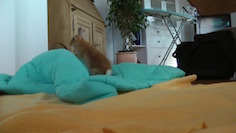}
           \includegraphics[height=1.7cm]{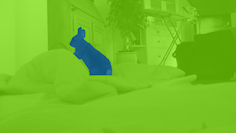}
           \caption{\small rabbits01}
           \label{fig:mouse}
       \end{subfigure}
     \caption{\small \textbf{FBMS-59}, five video frames of the training data set}\label{fig:animals}
 \end{figure}
   
 \begin{figure}
      \centering
      \begin{subfigure}[t] {0.16\textwidth}
      	 \centering
          \includegraphics[height=1.7cm]{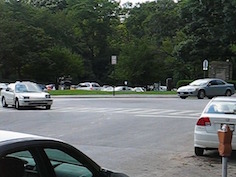}
          \includegraphics[height=1.7cm]{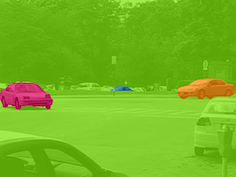}
          \caption{\small cars10}
          \label{fig:gull}
      \end{subfigure}
      ~ %add desired spacing between images, e. g. ~, \quad, \qquad, \hfill etc. 
      \begin{subfigure}[t] {0.18\textwidth}
       \centering
          \includegraphics[height=1.7cm]{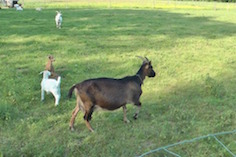}
          \includegraphics[height=1.7cm]{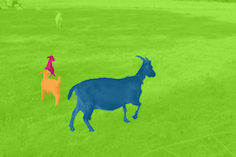}
          \caption{\small goats01}
          \label{fig:tiger}
      \end{subfigure}
      ~ %add desired spacing between images, e. g. ~, \quad, \qquad, \hfill etc. 
      \begin{subfigure}[t] {0.16\textwidth}
      	 \centering
          \includegraphics[height=1.7cm]{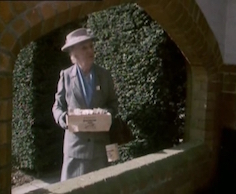}
          \includegraphics[height=1.7cm]{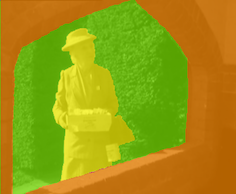}
          \caption{\small marple2}
          \label{fig:mouse}
      \end{subfigure}
      \begin{subfigure}[t] {0.19\textwidth}
      	  \centering
           \includegraphics[height=1.7cm]{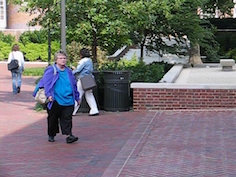}
           \includegraphics[height=1.7cm]{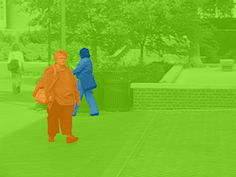}
           \caption{\small people2}
           \label{fig:mouse}
       \end{subfigure}
       \begin{subfigure}[t] {0.19\textwidth}
       	   \centering
            \includegraphics[height=1.7cm]{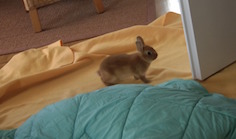}
            \includegraphics[height=1.7cm]{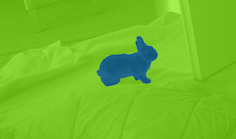}
            \caption{\small rabbits03}
            \label{fig:mouse}
        \end{subfigure}
      \caption{\small \textbf{FBMS-59}, five video frames of the test data set}\label{fig:animals}
  \end{figure} 
 % \end{comment}
  \vspace{2ex}
  \textbf{Complex Background Dataset \cite{narayana2013coherent}}
  The complex background data set contains five video sequences. A ground truth is provided for every fifth frame. A frame is always segmented into two segments, static background and freely moving objects. The ground truth does not provide multiple segments for different freely moving objects. This data set focuses especially on the difficulty of having a high variance in depth, such as a forest sequence. Trees grow at every depth level. Other videos like traffic show large objects which dominate the foreground, so that we have a high change in depth.
  
This data set can be viewed and downloaded at
\url{http://vis-www.cs.umass.edu/~narayana/motionsegmentation.html}
  %\begin{comment}
  \begin{figure}
        \centering
        \begin{subfigure}[t] {0.16\textwidth}
        	 \centering
            \includegraphics[height=1.7cm]{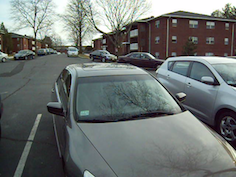}
            \includegraphics[height=1.7cm]{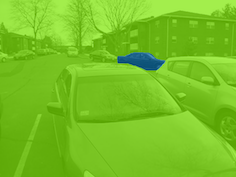}
            \caption{\small drive}
            \label{fig:gull}
        \end{subfigure}
        ~ %add desired spacing between images, e. g. ~, \quad, \qquad, \hfill etc. 
        \begin{subfigure}[t] {0.16\textwidth}
         \centering
            \includegraphics[height=1.7cm]{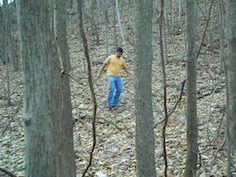}
            \includegraphics[height=1.7cm]{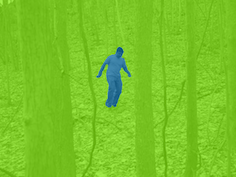}
            \caption{\small forest}
            \label{fig:tiger}
        \end{subfigure}
        ~ %add desired spacing between images, e. g. ~, \quad, \qquad, \hfill etc. 
        \begin{subfigure}[t] {0.17\textwidth}
        	 \centering
            \includegraphics[height=1.7cm]{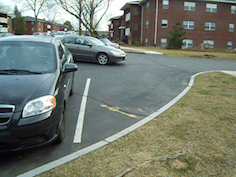}
            \includegraphics[height=1.7cm]{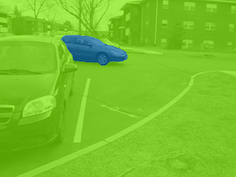}
            \caption{\small parking}
            \label{fig:mouse}
        \end{subfigure}
        \begin{subfigure}[t] {0.19\textwidth}
        	  \centering
             \includegraphics[height=1.7cm]{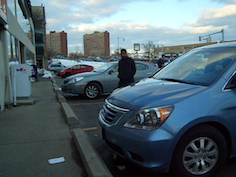}
             \includegraphics[height=1.7cm]{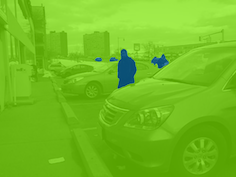}
             \caption{\small store}
             \label{fig:mouse}
         \end{subfigure}
         \begin{subfigure}[t] {0.17\textwidth}
         	   \centering
              \includegraphics[height=1.7cm]{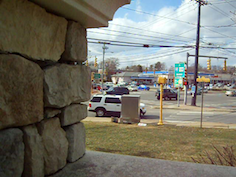}
              \includegraphics[height=1.7cm]{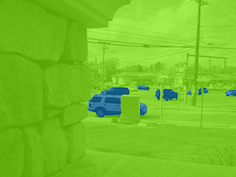}
              \caption{\small traffic}
              \label{fig:mouse}
          \end{subfigure}
        \caption{\small \textbf{Complex Background Dataset}}\label{fig:animals}
    \end{figure} 
  %\end{comment}
  
  \vspace{2ex}
  \textbf{Camouflaged Animals Dataset \cite{BideauL16}}
  The camouflaged animals data set includes nine video sequences extracted from YouTube videos \cite{utube_camouflage_1,utube_camouflage_2,utube_camouflage_3, utube_phasmid} and a ground truth for every fifth frame for each video. The ground truth contains multiple segments, the static background and one segment for every independently moving object. This data set addresses the problem of pure motion segmentation. It shows well-camouflaged animals in different surroundings. They are hard to detect in a static image, because of their excellent adaptation to their environment. In these cases, appearance is a weak feature which is not sufficient to detect camouflaged animals. The goal of this data set is to highlight the importance of motion as a feature for detecting objects that move differently from the camera.
  
  This data set can be viewed and downloaded at \url{http://vis-www.cs.umass.edu/motionSegmentation/}
  %\begin{comment}
  \begin{figure}
          \centering
          \begin{subfigure}[t] {0.16\textwidth}
          	 \centering
              \includegraphics[height=1.4cm]{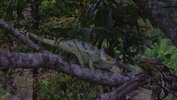}
              \includegraphics[height=1.4cm]{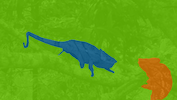}
              \caption{\small chameleon}
              \label{fig:gull}
          \end{subfigure}
          ~ %add desired spacing between images, e. g. ~, \quad, \qquad, \hfill etc. 
          \begin{subfigure}[t] {0.16\textwidth}
           \centering
              \includegraphics[height=1.4cm]{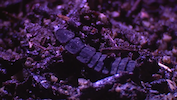}
              \includegraphics[height=1.4cm]{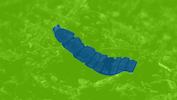}
              \caption{\small gl.-beetle}
              \label{fig:tiger}
          \end{subfigure}
          ~ %add desired spacing between images, e. g. ~, \quad, \qquad, \hfill etc. 
          \begin{subfigure}[t] {0.18\textwidth}
          	 \centering
              \includegraphics[height=1.4cm]{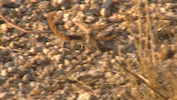}
              \includegraphics[height=1.4cm]{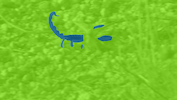}
              \caption{\small scorpion1}
              \label{fig:mouse}
          \end{subfigure}
          \begin{subfigure}[t] {0.18\textwidth}
          	  \centering
               \includegraphics[height=1.4cm]{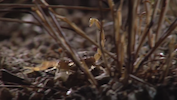}
               \includegraphics[height=1.4cm]{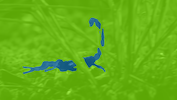}
               \caption{\small scorpion2}
               \label{fig:mouse}
           \end{subfigure}
           \begin{subfigure}[t] {0.18\textwidth}
           	   \centering
                \includegraphics[height=1.4cm]{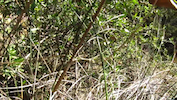}
                \includegraphics[height=1.4cm]{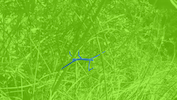}
                \caption{\small stickinsect}
                \label{fig:mouse}
            \end{subfigure}
          \caption{\small \textbf{Camouflaged Animals Dataset}}\label{fig:animals}
      \end{figure} 
   %\end{comment}   
  
\begin{figure}
            \centering
            \begin{subfigure}[t] {0.16\textwidth}
            	 \centering
                \includegraphics[height=1.4cm]{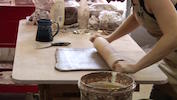}
                \includegraphics[height=1.4cm]{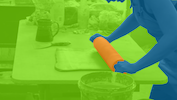}
                \includegraphics[height=1.4cm]{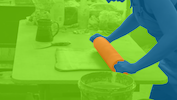}
                \includegraphics[height=1.4cm]{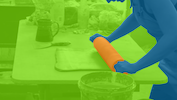}
                \includegraphics[height=1.4cm]{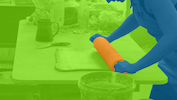}
                \caption{\small rolling pin}
                \label{fig:gull}
            \end{subfigure}
            ~ %add desired spacing between images, e. g. ~, \quad, \qquad, \hfill etc. 
            \begin{subfigure}[t] {0.16\textwidth}
             \centering
                \includegraphics[height=1.4cm]{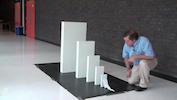}
                \includegraphics[height=1.4cm]{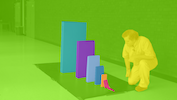}
                \includegraphics[height=1.4cm]{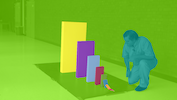}
                \includegraphics[height=1.4cm]{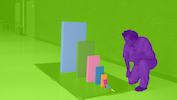}
                \includegraphics[height=1.4cm]{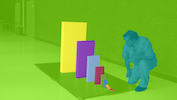}
                \caption{\small dominoes}
                \label{fig:tiger}
            \end{subfigure}
            ~ %add desired spacing between images, e. g. ~, \quad, \qquad, \hfill etc. 
            \begin{subfigure}[t] {0.18\textwidth}
            	 \centering
                \includegraphics[height=1.4cm]{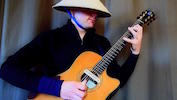}
                \includegraphics[height=1.4cm]{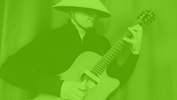}
                \includegraphics[height=1.4cm]{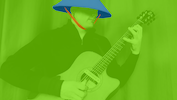}
                \includegraphics[height=1.4cm]{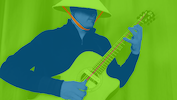}
                \includegraphics[height=1.4cm]{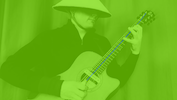}
                \caption{\small guitar}
                \label{fig:mouse}
            \end{subfigure}
            \begin{subfigure}[t] {0.18\textwidth}
            	  \centering
                 \includegraphics[height=1.4cm]{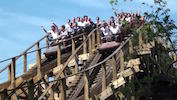}
                 \includegraphics[height=1.4cm]{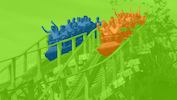}
                 \includegraphics[height=1.4cm]{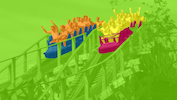}
                 \includegraphics[height=1.4cm]{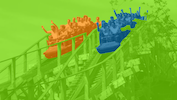}
                 \includegraphics[height=1.4cm]{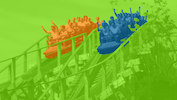}
                 \caption{\small roller coaster}
                 \label{fig:mouse}
             \end{subfigure}
             \begin{subfigure}[t] {0.18\textwidth}
             	   \centering
                  \includegraphics[height=1.4cm]{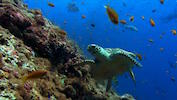}
                  \includegraphics[height=1.4cm]{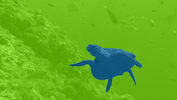}
                  \includegraphics[height=1.4cm]{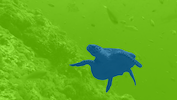}
                  \includegraphics[height=1.4cm]{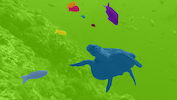}
                  \includegraphics[height=1.4cm]{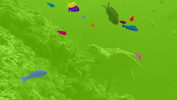}
                  \caption{\small see turtle}
                  \label{fig:mouse}
              \end{subfigure}
            \caption{\small \textbf{VSB100 motion subtask}, five video frames of the training set}\label{fig:VSB100train}
        \end{figure}

\begin{figure}
                    \centering
                    \begin{subfigure}[t] {0.16\textwidth}
                    	 \centering
                        \includegraphics[height=1.4cm]{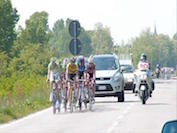}
                        \includegraphics[height=1.4cm]{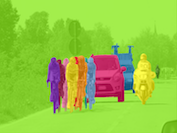}
                        \includegraphics[height=1.4cm]{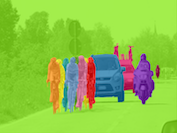}
                        \includegraphics[height=1.4cm]{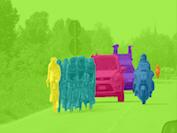}
                        \includegraphics[height=1.4cm]{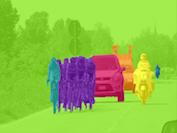}
                        \caption{\small bicycle race}
                        \label{fig:gull}
                    \end{subfigure}
                    ~ %add desired spacing between images, e. g. ~, \quad, \qquad, \hfill etc. 
                    \begin{subfigure}[t] {0.16\textwidth}
                     \centering
                        \includegraphics[height=1.4cm]{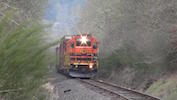}
                        \includegraphics[height=1.4cm]{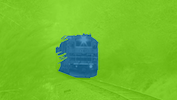}
                        \includegraphics[height=1.4cm]{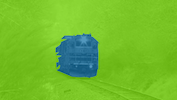}
                        \includegraphics[height=1.4cm]{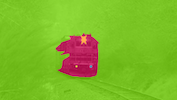}
                        \includegraphics[height=1.4cm]{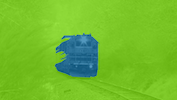}
                        \caption{\small freight train}
                        \label{fig:tiger}
                    \end{subfigure}
                    ~ %add desired spacing between images, e. g. ~, \quad, \qquad, \hfill etc. 
                    \begin{subfigure}[t] {0.18\textwidth}
                    	 \centering
                        \includegraphics[height=1.4cm]{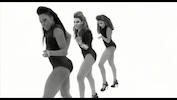}
                        \includegraphics[height=1.4cm]{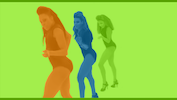}
                        \includegraphics[height=1.4cm]{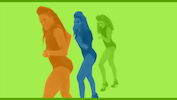}
                        \includegraphics[height=1.4cm]{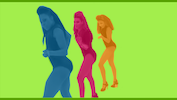}
                        \includegraphics[height=1.4cm]{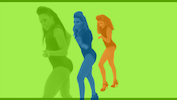}
                        \caption{\small beyonce}
                        \label{fig:mouse}
                    \end{subfigure}
                    \begin{subfigure}[t] {0.18\textwidth}
                    	  \centering
                         \includegraphics[height=1.4cm]{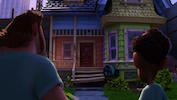}
                         \includegraphics[height=1.4cm]{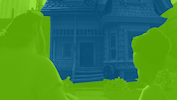}
                         \includegraphics[height=1.4cm]{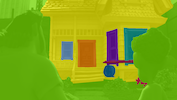}
                         \includegraphics[height=1.4cm]{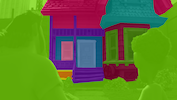}
                         \includegraphics[height=1.4cm]{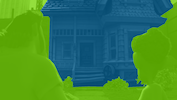}
                         \caption{\small up trailer}
                         \label{fig:mouse}
                     \end{subfigure}
                     \begin{subfigure}[t] {0.18\textwidth}
                     	   \centering
                          \includegraphics[height=1.4cm]{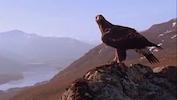}
                          \includegraphics[height=1.4cm]{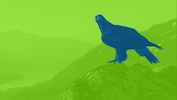}
                          \includegraphics[height=1.4cm]{fig/datasets/VSB100_Test/e_02.png}
                          \includegraphics[height=1.4cm]{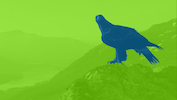}
                          \includegraphics[height=1.4cm]{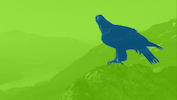}
                          \caption{\small planet earth 1}
                          \label{fig:mouse}
                      \end{subfigure}
                    \caption{\small \textbf{VSB100 motion subtask}, five video frames of the test set}\label{fig:VSB100test}
                \end{figure}

  \vspace{2ex}
  \textbf{VSB100 (motion subtask) \cite{galasso2013unified}} 
  VSB100 is a general video segmentation data set. It contains a lot of different videos, sequences of the animated video "up", short sequence of  a music video "beyonce" and a lot of natural scenes showing sports activities or animals. Not every scene shows a \textit{dominant object} or \textit{moving object}. Every 20th frame is annotated.
  Four different human annotations are provided, to measure the natural level of ambiguity of video segmentation. Multiple persons were asked to segment the video - objects, persons, animals, image parts etc. that describe the video sequence. More details for segmentation were not provided. Such that the goal of the segmentation is not defined precisely and the manually annotation shows the expected high ambiguity. 
  %\begin{comment}
  \begin{wrapfigure}{r}{0.3\textwidth}
  \vspace{-5ex}
          \begin{center}
            \includegraphics[width=0.26\textwidth]{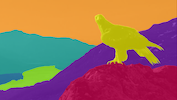}
            \includegraphics[width=0.26\textwidth]{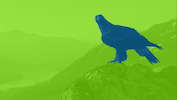}
          \end{center}
          \caption{\small From the video segmentation task to the motion subtask.}\label{fig:VSB100motionSubtask}
  \vspace{-2ex}
   \end{wrapfigure}
   %\end{comment}
  Additionally to the general video segmentation task, VSB100 provides a motion subtask (Figure \ref{fig:VSB100test} and Figure \ref{fig:VSB100train}). For the motion subtask, non-moving objects are ignored in the evaluation (Figure \ref{fig:VSB100motionSubtask}). Details regarding the motion segmentation problem are not provided. Questions like how are independently moving objects segmented? When do they form an own label, when are segments labeled together? Are moving objects segmented even if they are not moving? are not addressed. Addressing these questions are critical to solve the motion segmentation problem. 
  This data set can be viewed and downloaded at \url{http://lmb.informatik.uni-freiburg.de/resources/datasets/vsb.en.html}
  %\begin{comment}
        
  %\end{comment}
  \vspace{2ex}
  \textbf{Densly Annotated VIdeo Segmentation (Davis) \cite{Perazzi2016}} This data set consists of 50 high quality, Full HD video sequences. Every frame is annotated. Beside a new accuracy measurement this data set comes with a very detailed analysis of challenges and attributes covered in each video such as appearance changes, camera shake, dynamic background etc. The data sets goal is to segment the \textit{most dominant object}. Since all videos clearly show a dominant object it's pretty clear which object is the dominant object. However a lot of general videos, film sequences, sequences of a series, traffic sequences do not show a clear dominant object. In those situations it's not quite clear what is understood as the most dominant object. 
  The provided ground truth of the data set shows the most dominant object. Thus the ground truth always has two labels - one for the dominant object and one for background.
  Despite the fact, that just a binary segmentation is provided, they describe how to evaluate approaches generating multiple segments per frame in an accurate manner. It is suggested to pick the region with maximum region similarity to select the most similar object to the target object \cite{Perazzi2016}.
  This makes it possible to evaluate also more complex motion segmentation approaches, which segments all motion components in a frame, on this DAVIS data set.  
  %\begin{comment}
  \begin{figure}
            \centering
            \begin{subfigure}[t] {0.16\textwidth}
            	 \centering
                \includegraphics[height=1.4cm]{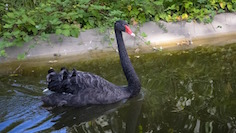}
                \includegraphics[height=1.4cm]{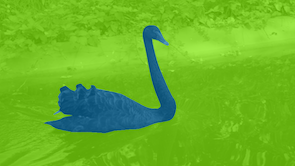}
                \caption{\small blackswan}
                \label{fig:gull}
            \end{subfigure}
            ~ %add desired spacing between images, e. g. ~, \quad, \qquad, \hfill etc. 
            \begin{subfigure}[t] {0.16\textwidth}
             \centering
                \includegraphics[height=1.4cm]{fig/datasets/DAVIS/breakdance.jpg}
                \includegraphics[height=1.4cm]{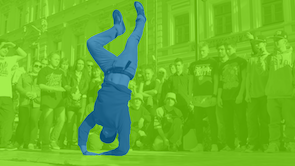}
                \caption{\small breakdance}
                \label{fig:tiger}
            \end{subfigure}
            ~ %add desired spacing between images, e. g. ~, \quad, \qquad, \hfill etc. 
            \begin{subfigure}[t] {0.18\textwidth}
            	 \centering
                \includegraphics[height=1.4cm]{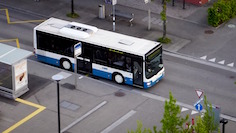}
                \includegraphics[height=1.4cm]{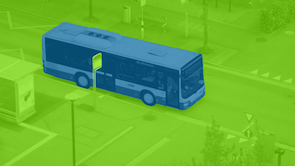}
                \caption{\small bus}
                \label{fig:mouse}
            \end{subfigure}
            \begin{subfigure}[t] {0.18\textwidth}
            	  \centering
                 \includegraphics[height=1.4cm]{fig/datasets/DAVIS/goat.jpg}
                 \includegraphics[height=1.4cm]{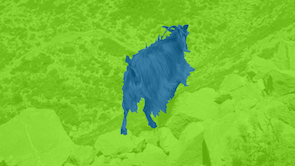}
                 \caption{\small goat}
                 \label{fig:mouse}
             \end{subfigure}
             \begin{subfigure}[t] {0.18\textwidth}
             	   \centering
                  \includegraphics[height=1.4cm]{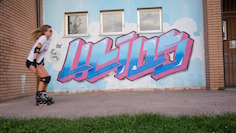}
                  \includegraphics[height=1.4cm]{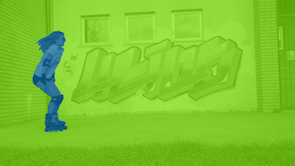}
                  \caption{\small rollerblade}
                  \label{fig:mouse}
              \end{subfigure}
            \caption{\small \textbf{Densly Annotated VIdeo Segmentation (Davis)}}\label{fig:animals}
        \end{figure} 
        
  %\end{comment}      
  \vspace{2ex}
  \textbf{SegTrack v2 \cite{li2013video}} is a small video segmentation data set, which comprises a 14 video sequences. This data set is an extension of the SegTrack Dataset \cite{tsai2012motion}.
  Compared to the SegTrack data set SegTrack v2 provides multiple object segments. Each frame is annotated. The image quality is not any more representative for current computer vision problems. Frame size varies between 640x360 pixel and 320x240 pixel. Its goal is to segment the \textit{object of interest}. The number of objects of interest is not limited. 
  
  This data set can be viewed an downloaded at
  \url{http://web.engr.oregonstate.edu/~lif/SegTrack2/index.html}
  %\begin{comment}
  \begin{figure}
              \centering
              \begin{subfigure}[t] {0.17\textwidth}
              	 \centering
                  \includegraphics[height=1.4cm]{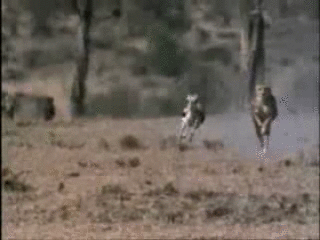}
                  \includegraphics[height=1.4cm]{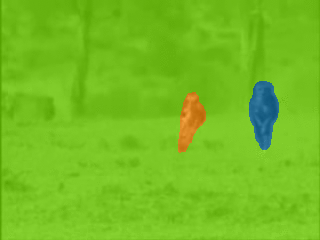}
                  \caption{\small chasedeer}
                  \label{fig:gull}
              \end{subfigure}
              ~ %add desired spacing between images, e. g. ~, \quad, \qquad, \hfill etc. 
              \begin{subfigure}[t] {0.16\textwidth}
               \centering
                  \includegraphics[height=1.4cm]{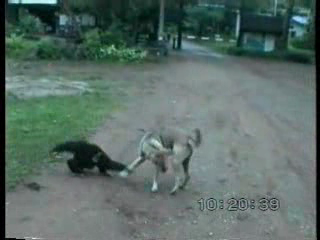}
                  \includegraphics[height=1.4cm]{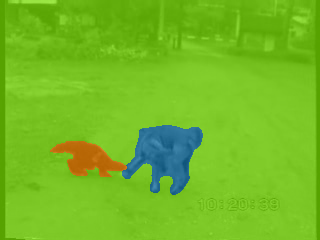}
                  \caption{\small monkeydog}
                  \label{fig:tiger}
              \end{subfigure}
              ~ %add desired spacing between images, e. g. ~, \quad, \qquad, \hfill etc. 
              \begin{subfigure}[t] {0.16\textwidth}
              	 \centering
                  \includegraphics[height=1.4cm]{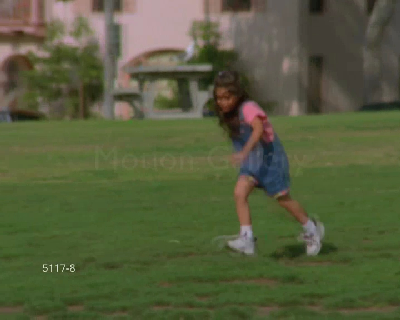}
                  \includegraphics[height=1.4cm]{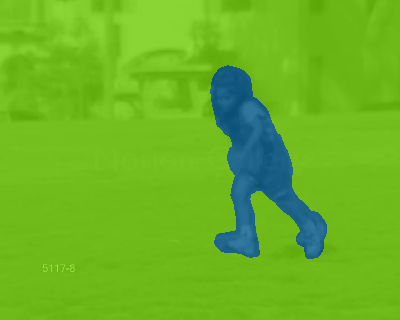}
                  \caption{\small girl}
                  \label{fig:mouse}
              \end{subfigure}
              \begin{subfigure}[t] {0.18\textwidth}
              	  \centering
                   \includegraphics[height=1.4cm]{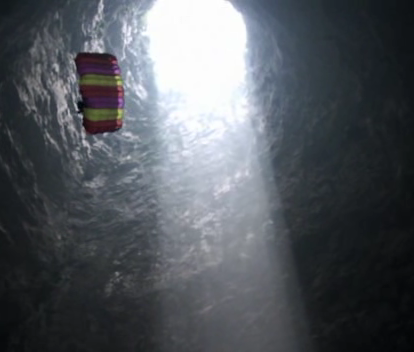}
                   \includegraphics[height=1.4cm]{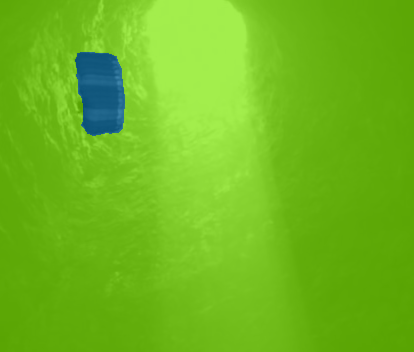}
                   \caption{\small parachute}
                   \label{fig:mouse}
               \end{subfigure}
               \begin{subfigure}[t] {0.19\textwidth}
               	   \centering
                    \includegraphics[height=1.4cm]{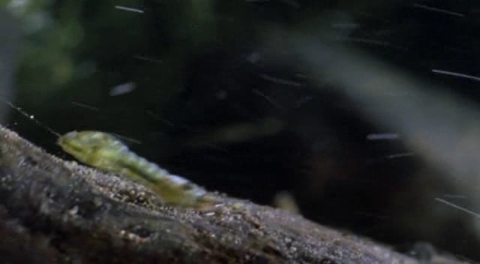}
                    \includegraphics[height=1.4cm]{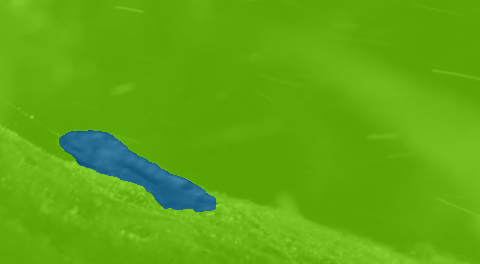}
                    \caption{\small worm}
                    \label{fig:mouse}
                \end{subfigure}
              \caption{\small \textbf{SegTrack v2}}\label{fig:animals}
          \end{figure} 
   %\end{comment}
  \vspace{2ex}
  
  All those data sets are addressing the video segmentation problem. However their goal is often not well defined. The idea of segmentation always comes with a criterion like motion, color, object etc. The motion segmentation problem alone is quite ambiguous like \cite{galasso2013unified} and the provided data set shows. Therefor a clear definition of the criterion is essential for segmentation as well as evaluation. 
  
  In the following sections we'll address the video segmentation problem under the aspect of motion segmentation. We start with a binary motion segmentation approach, extend this to the motion segmentation approach, which distinguishes between different independently moving objects. We conclude with prospects based on the fundamental motion segmentation problem which can finally lead to a better object understanding.
  
  \begin{table}  
  \begin{center}
  \scriptsize
  \rowcolors{2}{gray!25}{white}
    \begin{tabular}{p{2.9cm}p{2.2cm}cp{1.3cm}cc}
      \rowcolor{gray!50}
      Data Set & Goal & videos & size \newline (in px) & $ \frac{\mathrm{annotated \ frames}}{\mathrm{total \ frames}}$ & multiple labels \\
      FBMS-59 & motion segm. & 59 & $\sim$640x480& 0.05 & \ding{51} \\
      Complex Background & motion segm. & 5 & 640x480& 0.2 & \\
      Camouflaged Animals & motion segm. & 9 & 640x360 & 0.2 & \ding{51}\\
      VSB100 & video segm. with motion subtask & 100 & $\sim$1280x720 & 0.05 & \ding{51}\\
      Davis & most dominant object segm. & 50 & 854x480, 1920x1080 & 1 & \\
      SegTrack v2 & object of interest segm. & 14 & $\sim$320x240 & 1 & \ding{51}
    \end{tabular}
    \caption {\small \textbf{Overview}, the six reviewed video segmentation data sets}
  \end{center}
  \end{table}
  
  \section{Motion Segmentation}
  Image segmentation is a general problem in the area of computer vision and machine learning. The overall goal is to produce $k$ connected regions by pooling pixels, which share one or multiple common criteria. Those criteria might be color, texture or motion for example. Motion segmentation groups pixels which share the same motion. 
  
  3D motion or the displacement of a point in 3D, is observed as a pixel displacement in 2D on the image plane. However it is hard to distinguish for tiny motions whether they are moving or not. Motion magnification makes it possible for example to magnify tiny motions such as a bridge wiggling in the wind such that even 'invisible' motion become visible. The boarder between stationary objects and moving objects is fluent.
  If a person is walking there is mostly a short time period where one foot is moving, but the other foot stands still. Do we want to segment just the part of the person, which is moving or the entire moving person? Those difficulties \textit{(1) are those pixels moving or not?} and \textit{(2) Is just part of the object moving or the entire object?} turns creating a ground truth into a challenging problem.
  
  Since the criterion "motion" alone is hard to evaluate we will refer to object motion instead of the motion of a single pixel.
  This is a useful and practical simplification. Using this simplification the entire object needs to be segmented even if just part of it is moving. If just one foot of a walking person is moving, we'll segment the entire object or if just the shovel of a digger is moving, we'll segment the entire digger - not the shovel only.
  
  In the following we'll address the problem of motion segmentation as a binary segmentation problem and extend this task to the more general task of motion segmentation with a flexible amount of motion components. 
  
  \subsection{Definition: Binary Motion Segmentation}
  %\begin{comment}
  \begin{SCfigure}[1.6][h]
  \includegraphics[width=0.25\linewidth]{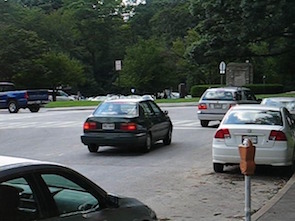}
     \includegraphics[width=0.25\linewidth]{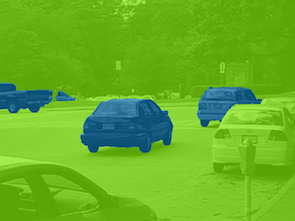}
     \caption{\textbf{Two Labels} Every pixel is given one of two labels: static environment or moving objects. Left to Right: original image, correct binary segmentation}
  \label{fig:phasmid_plain}
  \end{SCfigure}
  %\end{comment}
  Binary motion segmentation segments each video frame into two components. It is  distinguished between (1) static environment and (2) independently moving objects moving differently than the camera motion. The environment itself is not moving, however the pixels describing static environment can show a displacement from frame $t$ to $t+1$ due accordingly to the camera motion. We define motion segmentation as follows:
  
\begin{compactenum}[(I)]
\item Every pixel is given one of {\bf two labels}: static environment or moving objects.
\item If only part of an object is moving (like a moving person with a
  stationary foot), the {\bf entire object} should be segmented.
\item {\bf All freely moving objects} (not just one) should
be segmented, but nothing else. We do not considered tethered objects such 
as trees to be freely moving.
\item Stationary objects are not segmented, even when
  they moved before or will move in the future. We consider 
  segmentation of previously moving objects to be {\em
    tracking}. Our focus is on segmentation by motion analysis.
\end{compactenum}
%\begin{comment}
\begin{SCfigure}[1.6][h]
\includegraphics[width=0.25\linewidth]{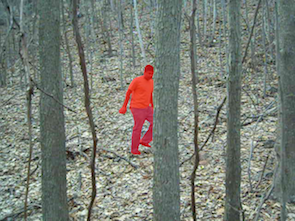}
   \includegraphics[width=0.25\linewidth]{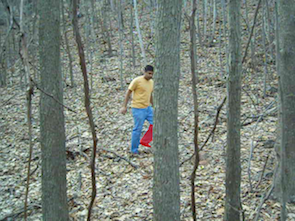}
   \caption{\textbf{Entire Object} If only part of an object is moving (here a moving person with a
     stationary foot), the entire object should be segmented. Left to Right: correct segmentation, wrong segmentation.}
\label{fig:phasmid_plain}
\end{SCfigure}

\begin{SCfigure}[1.6][h]
\includegraphics[width=0.25\linewidth]{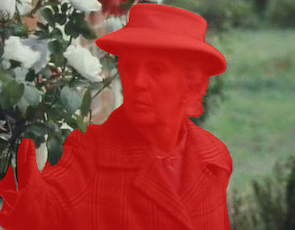}
   \includegraphics[width=0.25\linewidth]{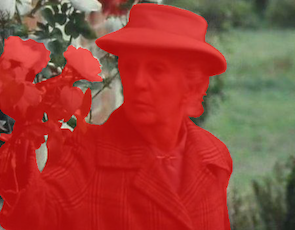}
   \caption{\textbf{All Freely Moving Objects} (not just one) should
   be segmented, but nothing else. We do not considered tethered objects such as this rose bush to be freely moving. Left to Right: correct segmentation, wrong segmentation.}
\label{fig:phasmid_plain}
\end{SCfigure}
 %\end{comment} 
  \subsection{Definition: Motion Segmentation}
	\label{sec:MoSegDef}
\begin{SCfigure}[][h!]
         \includegraphics[width=0.25\linewidth]{fig/cars2_01.jpg}
            \includegraphics[width=0.25\linewidth]{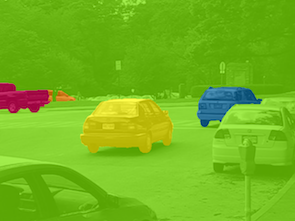}
            \caption{\textbf{$k$ Labels} Every pixel is given one of $k$ labels: $k$ is the number of independently moving objects and static environment. Left to right: original image, motion segmentation}
         \label{fig:4cars}
       \end{SCfigure}

   It is critical to be aware that motion Segmentation, is a segmentation task which groups pixels which share the same motion.
   Just in sense of motion it's often not possible to distinguish between different objects, since they share the same motion. Since they share the same motion they form a common pixel group which is segmented together. 
   Segmenting a video into $k$ independently moving objects is clear if the moving objects a not touching in 3D and clearly move independently like the four cars shown in Figure \ref{fig:4cars}. However there are a few cases where segmenting a frame into $k$ independently moving objects is challenging: \textit{(1) two persons walking hand in hand, (2) a person jumping onto a carriage, which is pulled by a horse or (3) laundry fluttering in the wind}. After defining motion segmentation as a general segmentation problem of a video into $k$ moving objects, we'll take a closer look at those boarder cases. 
  %\begin{comment} 
  % \end{comment}    
   We build upon our previous definition of binary motion segmentation:
  
  \begin{compactenum}[(I)]
  \item Every pixel is given one of {\bf $k$ labels}. $k$ is the number of observed independently moving objects and static environment.
  \item If only part of an object is moving (like a moving person with a
    stationary foot), the {\bf entire object} should be segmented.
  \item {\bf All freely moving objects} (not just one) should
  be segmented, but nothing else. We do not considered tethered objects such as trees to be freely moving. 
  \item Stationary objects are not segmented, even when
    they moved before or will move in the future. We consider 
    segmentation of previously moving objects to be {\em
      tracking}. Our focus is on segmentation by motion analysis.
  \item If objects are moving together and are connected in 3D, they should be segmented together since they share the same motion. A person carrying a basket - here the person and the basket are forming one common motion component.
  \item If objects are connected in 3D but move independently from each other, they should be segmented separately since they do not share the same motion. A person walking with a leashed dog. The person and the dog are connected in 3D, but move independently from each other. Thus the person and the dog get their own motion segment.
  \item An object which is not moving (but could be connected in 3D with an other moving object) should be not segmented unless it is considered an integral part of an other object. Is a person sitting on a stationary chair, then the chair should not be segmented. However if the chair is moving with that person (for example a wheelchair), then the person should be segmented together with the chair following rule (V).
 % \begin{comment}
  \begin{SCfigure}[][h]
      \includegraphics[width=0.25\linewidth]{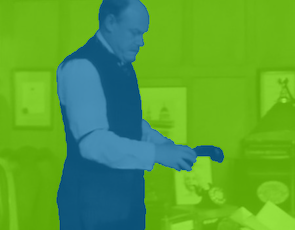}
     \includegraphics[width=0.25\linewidth]{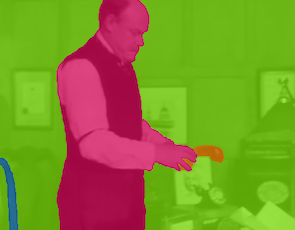}
         \caption{\textbf{Objects moving together} and connected in 3D like man and phone should be segmented together since they share the same motion. The chair however was moved before by the man, in this frame the chair is not moving anymore such that it should not get a separate motion component here. Left to Right: correct motion segmentation, wrong motion segmentation.}
      \label{fig:phasmid_plain}
 \end{SCfigure}
  %\end{comment}
  \end{compactenum}
\vspace{2ex}
  Based on the provided definition (I-VII) of motion segmentation, we discuss the three previously mentioned challenging cases.
  \begin{itemize}
  \item \textit{Two persons walking hand in hand} The two persons are connected in 3D and move pretty much independently even if there might be some influence from one person to the other. Basically there are two independent "motion sources" thus we observe two independent motions and label both persons separately according to (VI). This is a challenging case since as long the persons are connected in 3D it is hard to judge based on the video whether this is one complex motion component or two simpler and probably similar motion components.
  \item \textit{A person jumping onto a carriage, which is pulled by a horse} We start with carriage and horse. This situation corresponds to (V). The horse is pulling the carriage such that the carriage is moving with the horse. Carriage and horse form together one independently moving objects. Now we consider the man jumping onto the carriage. This situation gets significantly more tricky. The man is for sure moving independently at the beginning thus rule (VI) can be applied however in the later run if the man doesn't move significantly independent anymore - sitting on the carriage, the man can be considered to be segmented with the carriage and horse (V). This is a very difficult case and can be interpreted differently, however we consider together moving objects as one moving object, if they move as a whole independently.
  \item \textit{laundry fluttering in the wind.} Even though laundry does not belong to a tethered object such as the leaves of a tree, which a wiggling in the wind, we do not consider laundry as an independently moving object. This situation is quite similar as described in (III). Thus laundry fluttering in the wind is not considered to be a freely moving object.
  \end{itemize}

\subsubsection{Motion based Object Understanding}
  By looking on the motion segmentation problem from a higher level it becomes clear, that motion is a very informative criterion, regarding object understanding - understanding our environment we move and live in. The \textit{object} is a very abstract description of an entity. The concept of an \textit{object} depends of the observers perception and interest as well as the context.
  We might see an house as one object, but we could also become more detailed and recognize window, door and roof as independent objects. If I am driving a car I recognize a pedestrian with its bag, gloves, shoes etc. as one object however if this person goes into a fashion store all the clothes might be of interest and might be perceived as separate objects by the vendor. An \textit{object} is the perception of an entity. Subobjects form the simplified entire \textit{object} by putting all the subobjects into relation with each other. Motion is a property which is able to join subobjects to one entire object (def. V), but also to separate objects as soon as objects get disconnected in 3D and follow different motion patterns. Without any previous object knowledge I can step into a room perceiving a desk, with monitor, laptop, pencils and a book as a single very complex object. As soon as I move parts of the bigger object, like the book, I learn that the book is an independent object not connected to the desk. Without any previous knowledge about books, monitors or pencils I am able to understand that those are unknown independent objects. Motion, our own motion as well as the motion of objects, is incredible helpful for understanding our environment we move and live in. Such that a detailed motion analysis is critical for orientation in 3D
  and surviving in nature.
\begin{comment} 
  \section{FBMS-59: Improved Ground Truth}
  Following our definition of Motion Segmentation we worked through the existing FBMS-59 data set, the most popular motion segmentation data set. The provided ground truth showed several wrong segments, which clearly not describe moving objects such as walls, buildings etc. Furthermore objects which are moving were not segmented or segments which are not moving were segmented.
  Figure \ref{fig:newGT} shows an extract of the corrections we made. For details we recommend to watch the motion segmentation videos provided on our website.
  The ground truth for all video sequences corresponding to our provided definition are available at \url{http://people.cs.umass.edu/~pbideau/}.
  \end{comment}
  %\begin{comment}
  \begin{figure}
                  \centering
                  \begin{subfigure}[t] {0.24\textwidth}
                  	 \centering
                      \includegraphics[width=0.95\textwidth]{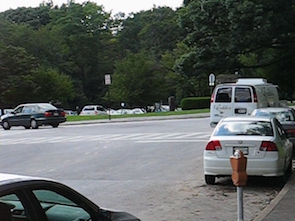}
                      \includegraphics[width=0.95\textwidth]{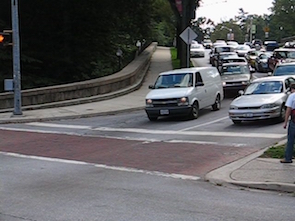}
                      \includegraphics[width=0.95\textwidth]{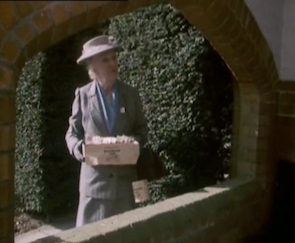}
                      \includegraphics[width=0.95\textwidth]{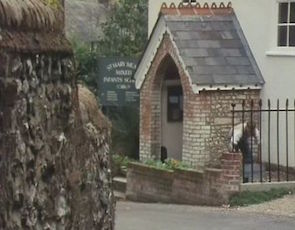}
                      \includegraphics[width=0.95\textwidth]{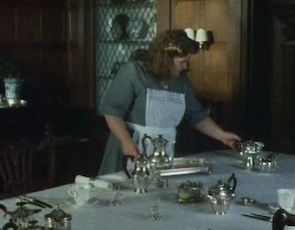}
                      \includegraphics[width=0.95\textwidth]{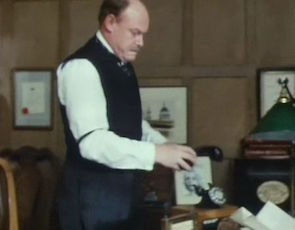}
                       \includegraphics[width=0.95\textwidth]{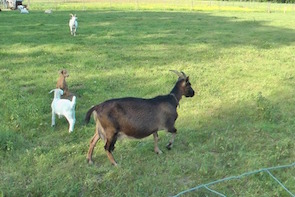}
                       \includegraphics[width=0.95\textwidth]{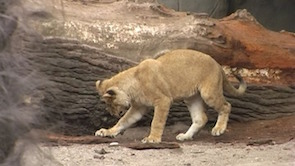}
                      \caption{\small original frame}
                      \label{fig:gull}
                  \end{subfigure}
                   \begin{subfigure}[t] {0.24\textwidth}
                      \centering
                      \includegraphics[width=0.95\textwidth]{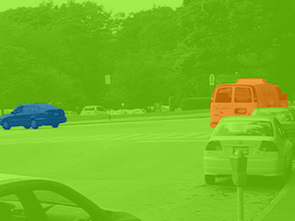}
                      \includegraphics[width=0.95\textwidth]{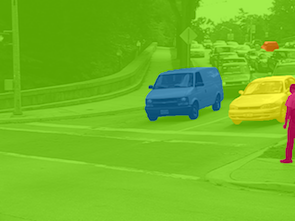}
                      \includegraphics[width=0.95\textwidth]{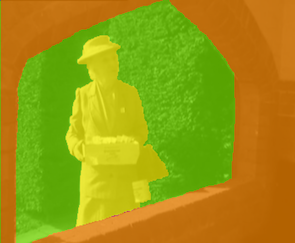}
                      \includegraphics[width=0.95\textwidth]{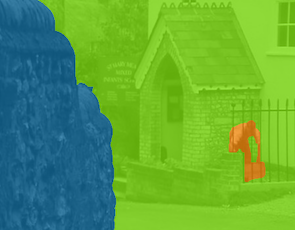}
                      \includegraphics[width=0.95\textwidth]{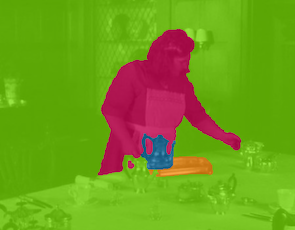}
                      \includegraphics[width=0.95\textwidth]{fig/new_gt/marple13_40_wrong.png}
                      \includegraphics[width=0.95\textwidth]{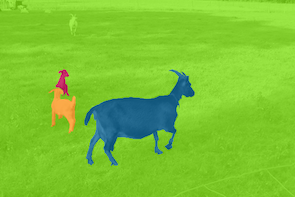}
                      \includegraphics[width=0.95\textwidth]{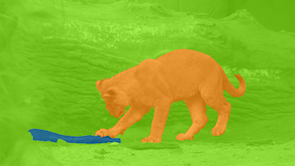}
                      \caption{\small FBMS-59}
                      \label{fig:gull}
                  \end{subfigure}
                  \begin{subfigure}[t] {0.24\textwidth}
                      \centering
                       \includegraphics[width=0.95\textwidth]{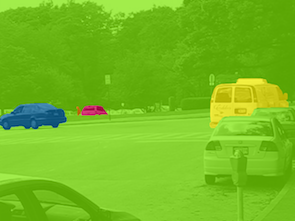}
                    \includegraphics[width=0.95\textwidth]{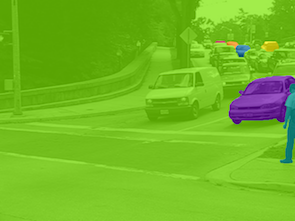}
                      \includegraphics[width=0.95\textwidth]{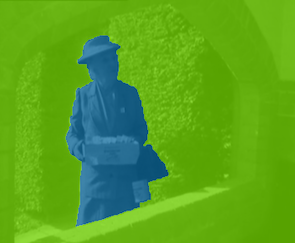}
                      \includegraphics[width=0.95\textwidth]{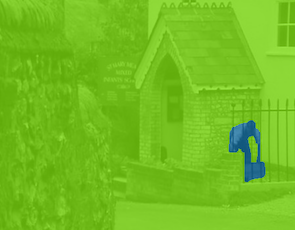}
                      \includegraphics[width=0.95\textwidth]{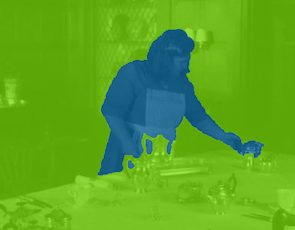}
                      \includegraphics[width=0.95\textwidth]{fig/new_gt/marple13_40_correct.png}
                      \includegraphics[width=0.95\textwidth]{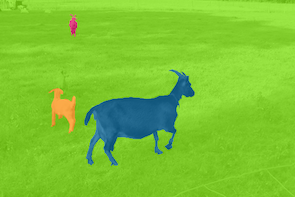}
                      \includegraphics[width=0.95\textwidth]{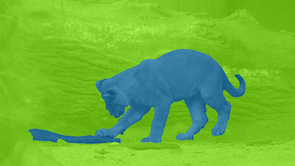}
                      \caption{\small new gt}
                      \label{fig:gull}
                  \end{subfigure}
               \caption{\small \textbf{new motion segmentation ground truth of FBMS-59 according to our motion segmentaiton definition}, 8 videos of the FBMS-59 data set. Top to Bottom: cars3, cars9, marple2, marple10, marple12, marple13, goats and lion01}\label{fig:newGT}
           \end{figure} 
            
  %\end{comment}          
            
 \section{Conclusion}
The ambiguity of video segmentation is a current problem in computer vision. There are several data sets addressing the video segmentation problem. However the lack of a precise problem definition leads to ambiguous segmentations and inaccurate evaluations. It is critical that the ground truth matches the addressed problem. It's not possible to evaluate object segmentation methods on a motion segmentation data set and the other way around and obtaining a meaningful evaluation. By  reviewing current data sets, providing a detailed definition of motion segmentation with several examples and a discussion of critical cases, where the definition of motion segmentation might still be not obvious at the first glance, we hope to clarify the video segmentation problem under the aspect of motion segmentation.
  %\section{Validation Methods}
  
\bibliographystyle{abbrv}
\bibliography{references}
  
  \end{document}